\documentclass[10pt,twocolumn,letterpaper]{article}

\usepackage{cvpr}
\usepackage{times}
\usepackage{epsfig}
\usepackage{graphicx}
\usepackage{amsmath}
\usepackage{amssymb}
\usepackage{subfigure}
\usepackage{multirow}
\usepackage{epsfig,epstopdf}
\usepackage{bbding}

\usepackage[pagebackref=true,breaklinks=true,letterpaper=true,colorlinks,bookmarks=false]{hyperref}

\cvprfinalcopy 

\def\cvprPaperID{1329} 

\ifcvprfinal\fi
\begin{document}

\title{Residual Dense Network for Image Super-Resolution}

\author{Yulun Zhang$^{1}$, Yapeng Tian$^{2}$, Yu Kong$^{1}$, Bineng Zhong$^{1}$, Yun Fu$^{1,3}$\\
$^{1}$Department of Electrical and Computer Engineering, Northeastern University, Boston, USA\\
$^{2}$Department of Computer Science, University of Rochester, Rochester, USA\\
$^{3}$College of Computer and Information Science, Northeastern University, Boston, USA\\
{\tt\small yulun100@gmail.com, yapengtian@rochester.edu, bnzhong@hqu.edu.cn, \{yukong,yunfu\}@ece.neu.edu}
}

\maketitle


\begin{abstract}
A very deep convolutional neural network (CNN) has recently achieved great success for image super-resolution (SR) and offered hierarchical features as well. However, most deep CNN based SR models do not make full use of the hierarchical features from the original low-resolution (LR) images, thereby achieving relatively-low performance. In this paper, we propose a novel residual dense network (RDN) to address this problem in image SR. We fully exploit the hierarchical features from all the convolutional layers. Specifically, we propose residual dense block (RDB) to extract abundant local features via dense connected convolutional layers. RDB further allows direct connections from the state of preceding RDB to all the layers of current RDB, leading to a contiguous memory (CM) mechanism. Local feature fusion in RDB is then used to adaptively learn more effective features from preceding and current local features and stabilizes the training of wider network. After fully obtaining dense local features, we use global feature fusion to jointly and adaptively learn global hierarchical features in a holistic way. Experiments on benchmark datasets with different degradation models show that our RDN achieves favorable performance against state-of-the-art methods.
\end{abstract}

\section{Introduction}
Single image Super-Resolution (SISR)  aims to generate a visually pleasing high-resolution (HR) image from its degraded low-resolution (LR) measurement. SISR is used in various computer vision tasks, such as security and surveillance imaging~\cite{zou2012very}, medical imaging~\cite{shi2013cardiac}, and image generation~\cite{karras2018progressive}. While image SR is an ill-posed inverse procedure, since there exists a multitude of solutions for any LR input. To tackle this inverse problem, plenty of image SR algorithms have been proposed, including interpolation-based~\cite{zhang2006edge}, reconstruction-based~\cite{zhang2012single}, and learning-based methods~\cite{timofte2013anchored,timofte2014a+,peleg2014statistical,dong2014learning,schulter2015fast,huang2015single,kim2016accurate,tong2017image,zhang2018learning}.
\begin{figure}[t]
\centering
\centerline{
\subfigure[{ Residual block}]{
\label{fig:blocks_RB}
\includegraphics[scale = 0.68]{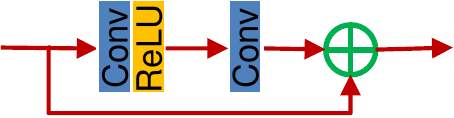}}
\subfigure[{ Dense block}]{
\label{fig:blocks_DB}
\includegraphics[scale = 0.68]{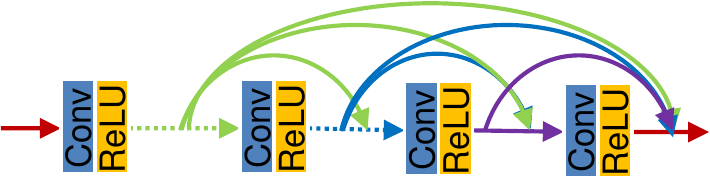}}
}
\centerline{
\subfigure[{ Residual dense block}]{
\label{fig:blocks_RDB}
\includegraphics[scale = 0.68]{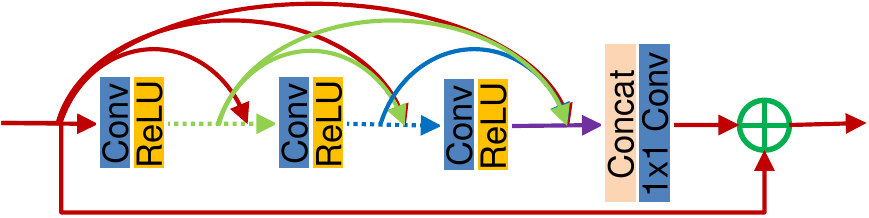}}
}

\caption{Comparison of prior network structures (a,b) and our residual dense block (c). (a) Residual block in MDSR~\cite{lim2017enhanced}. (b) Dense block in SRDenseNet~\cite{tong2017image}. (c) Our residual dense block.}  
\label{fig:blocks_RB_DB_RDB}
\vspace{-4mm}
\end{figure}

Among them, Dong~et al.~\cite{dong2014learning} firstly introduced a three-layer convolutional neural network (CNN) into image SR and achieved significant improvement over conventional methods. Kim et al. increased the network depth in VDSR~\cite{kim2016accurate} and DRCN~\cite{kim2016deeply} by using gradient clipping, skip connection, or recursive-supervision to ease the difficulty of training deep network. By using effective building modules, the networks for image SR are further made deeper and wider with better performance. Lim et al. used residual blocks (Fig.~\ref{fig:blocks_RB}) to build a very wide network EDSR~\cite{lim2017enhanced} with residual scaling~\cite{szegedy2017inception} and a very deep one MDSR~\cite{lim2017enhanced}. Tai et al. proposed memory block to build MemNet~\cite{tai2017memnet}. As the network depth grows, the features in each convolutional layer would be hierarchical with different receptive fields. However, these methods neglect to fully use information of each convolutional layer. Although the gate unit in memory block was proposed to control short-term memory~\cite{tai2017memnet}, the local convolutional layers don't have direct access to the subsequent layers. So it's hard to say memory block makes full use of the information from all the layers within it. 

Furthermore, objects in images have different scales, angles of view, and aspect ratios. Hierarchical features from a very deep network would give more clues for reconstruction. While, most deep learning (DL) based methods (e.g., VDSR~\cite{kim2016accurate}, LapSRN~\cite{lai2017deep}, and EDSR~\cite{lim2017enhanced}) neglect to use hierarchical features for reconstruction. Although memory block~\cite{tai2017memnet} also takes information from preceding memory blocks as input, the multi-level features are not extracted from the original LR image. MemNet interpolates the original LR image to the desired size to form the input. This pre-processing step not only increases computation complexity quadratically, but also loses some details of the original LR image. Tong et al. introduced dense block (Fig.~\ref{fig:blocks_DB}) for image SR with relatively low growth rate (e.g.,16). According to our experiments (see Section~\ref{subsec:study_DCG}), higher growth rate can further improve the performance of the network. While, it would be hard to train a wider network with dense blocks in Fig.~\ref{fig:blocks_DB}. 

\begin{figure*}[htbp]
\centering
\includegraphics[scale = 1]{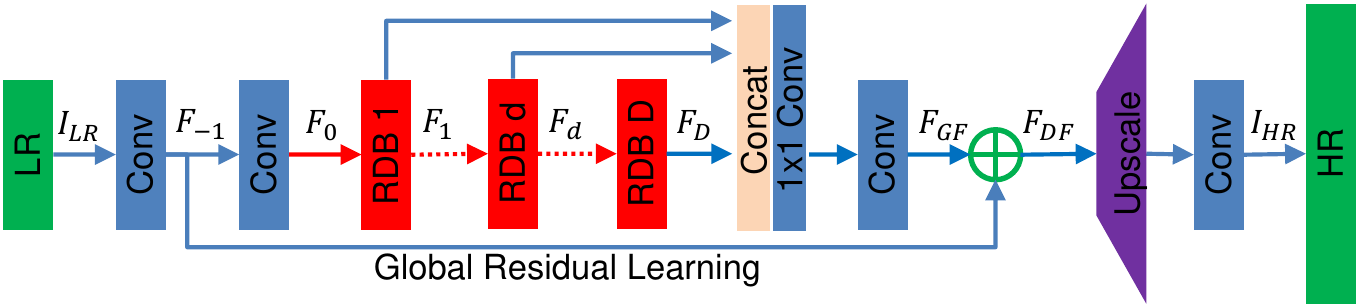}
\caption{The architecture of our proposed residual dense network (RDN).}
\label{fig:RDN} 
\vspace{-4mm}  
\end{figure*}

To address these drawbacks, we propose residual dense network (RDN) (Fig.~\ref{fig:RDN}) to fully make use of all the hierarchical features from the original LR image with our proposed residual dense block (Fig.~\ref{fig:blocks_RDB}). It's hard and impractical for a very deep network to directly extract the output of each convolutional layer in the LR space. We propose residual dense block (RDB) as the building module for RDN. RDB consists dense connected layers and local feature fusion (LFF) with local residual learning (LRL). Our RDB also support contiguous memory among RDBs. The output of one RDB has direct access to each layer of the next RDB, resulting in a contiguous state pass. Each convolutional layer in RDB has access to all the subsequent layers and passes on information that needs to be preserved~\cite{huang2017densely}. Concatenating the states of preceding RDB and all the preceding layers within the current RDB, LFF extracts local dense feature by adaptively preserving the information. Moreover, LFF allows very high growth rate by stabilizing the training of wider network. After extracting multi-level local dense features, we further conduct global feature fusion (GFF) to adaptively preserve the hierarchical features in a global way. As depicted in Figs.~\ref{fig:RDN} and~\ref{fig:RDB}, each layer has direct access to the original LR input, leading to an implicit deep supervision~\cite{lee2015deeply}.

In summary, our main contributions are three-fold:
\begin{itemize}
\item We propose a unified frame work residual dense network (RDN) for high-quality image SR with different degradation models. The network makes full use of all the hierarchical features from the original LR image.
\end{itemize}
\begin{itemize}
\item We propose residual dense block (RDB), which can not only read state from the preceding RDB via a contiguous memory (CM) mechanism, but also fully utilize all the layers within it via local dense connections. The accumulated features are then adaptively preserved by local feature fusion (LFF).
\end{itemize}
\begin{itemize}
\item We propose global feature fusion to adaptively fuse hierarchical features from all RDBs in the LR space. With global residual learning, we combine the shallow features and deep features together, resulting in global dense features from the original LR image. 
\end{itemize}

\section{Related Work}
Recently, deep learning (DL)-based methods have achieved dramatic advantages against conventional methods in computer vision~\cite{zhang2017image,zhang2018density,zhang2018densely,li2018tell}. Due to the limited space, we only discuss some works on image SR. Dong et al. proposed SRCNN~\cite{dong2014learning}, establishing an end-to-end mapping between the interpolated LR images and their HR counterparts for the first time. This baseline was then further improved mainly by increasing network depth or sharing network weights. VDSR~\cite{kim2016accurate} and IRCNN~\cite{zhang2017learning} increased the network depth by stacking more convolutional layers with residual learning. DRCN~\cite{kim2016deeply} firstly introduced recursive learning in a very deep network for parameter sharing. Tai et al. introduced recursive blocks in DRRN~\cite{tai2017image} and memory block in Memnet~\cite{tai2017memnet} for deeper networks. All of these methods need to interpolate the original LR images to the desired size before applying them into the networks. This pre-processing step not only increases computation complexity quadratically~\cite{dong2016accelerating}, but also over-smooths and blurs the original LR image, from which some details are lost. As a result, these methods extract features from the interpolated LR images, failing to establish an end-to-end mapping from the original LR to HR images. 

To solve the problem above, Dong et al.~\cite{dong2016accelerating} directly took the original LR image as input and introduced a transposed convolution layer (also known as deconvolution layer) for upsampling to the fine resolution. Shi et al. proposed ESPCN~\cite{shi2016real}, where an efficient sub-pixel convolution layer was introduced to upscale the final LR feature maps into the HR output. The efficient sub-pixel convolution layer was then adopted in SRResNet~\cite{ledig2017photo} and EDSR~\cite{lim2017enhanced}, which took advantage of residual leanrning~\cite{he2016deep}. All of these methods extracted features in the LR space and upscaled the final LR features with transposed or sub-pixel convolution layer. By doing so, these networks can either be capable of real-time SR (e.g., FSRCNN and ESPCN), or be built to be very deep/wide (e.g., SRResNet and EDSR). However, all of these methods stack building modules (e.g., Conv layer in FSRCNN, residual block in SRResNet and EDSR) in a chain way. They neglect to adequately utilize information from each Conv layer and only adopt CNN features from the last Conv layer in LR space for upscaling.

Recently, Huang et al. proposed DenseNet, which allows direct connections between any two layers within the same dense block~\cite{huang2017densely}. With the local dense connections, each layer reads information from all the preceding layers within the same dense block. The dense connection was introduced among memory blocks~\cite{tai2017memnet} and dense blocks~\cite{tong2017image}. More differences between DenseNet/SRDenseNet/MemNet and our RDN would be discussed in Section~\ref{sec:discussions}.

The aforementioned DL-based image SR methods have achieved significant improvement over conventional SR methods, but all of them lose some useful hierarchical features from the original LR image. Hierarchical features produced by a very deep network are useful for image restoration tasks (e.g., image SR). To fix this case, we propose residual dense network (RDN) to extract and adaptively fuse features from all the layers in the LR space efficiently. We will detail our RDN in next section.

\section{Residual Dense Network for Image SR}
\subsection{Network Structure}
\label{sec:network}
As shown in Fig.~\ref{fig:RDN}, our RDN mainly consists four parts: shallow feature extraction net (SFENet), redidual dense blocks (RDBs), dense feature fusion (DFF), and finally the up-sampling net (UPNet). Let's denote $I_{LR}$ and $I_{SR}$ as the input and output of RDN. Specifically, we use two Conv layers to extract shallow features. The first Conv layer extracts features $F_{-1}$ from the LR input.
\begin{align}
\begin{split}
\label{eq:SFE1}
F_{-1}=H_{SFE1}\left ( I_{LR} \right ),
\end{split}
\end{align}
where $H_{SFE1}\left ( \cdot  \right )$ denotes convolution operation. $F_{-1}$ is then used for further shallow feature extraction and global residual learning. So we can further have
\begin{align}
\begin{split}
\label{eq:SFE2}
F_{0}=H_{SFE2}\left ( F_{-1} \right ),
\end{split}
\end{align}
where $H_{SFE2}\left ( \cdot  \right )$ denotes convolution operation of the second shallow feature extraction layer and is used as input to residual dense blocks. Supposing we have $D$ residual dense blocks, the output $F_{d}$ of the $d$-th RDB can be obtained by
\begin{align}
\begin{split}
\label{eq:F_d_RDN}
F_{d}&=H_{RDB,d}\left ( F_{d-1} \right )\\
&=H_{RDB,d}\left ( H_{RDB,{d-1}}\left ( \cdots \left ( H_{RDB,1}\left ( F_{0} \right ) \right ) \cdots \right ) \right ),
\end{split}
\end{align}
where $H_{RDB,d}$ denotes the operations of the $d$-th RDB. $H_{RDB,d}$ can be a composite function of operations, such as convolution and rectified linear units (ReLU)~\cite{glorot2011deep}. As $F_{d}$ is produced by the $d$-th RDB fully utilizing each convolutional layers within the block, we can view $F_{d}$ as local feature. More details about RDB will be given in Section~\ref{subsec:RDB}.

After extracting hierarchical features with a set of RDBs, we further conduct dense feature fusion (DFF), which includes global feature fusion (GFF) and global residual learning (GRL). DFF makes full use of features from all the preceding layers and can be represented as
\begin{align}
\begin{split}
\label{eq:DFF}
F_{DF}=H_{DFF}\left ( F_{-1},F_{0},F_{1},\cdots ,F_{D} \right ),
\end{split}
\end{align}  
where $F_{DF}$ is the output feature-maps of DFF by utilizing a composite function $H_{DFF}$. More details about DFF will be shown in Section~\ref{subsec:DFF}. 

After extracting local and global features in the LR space, we stack a up-sampling net (UPNet) in the HR space. Inspired by~\cite{lim2017enhanced}, we utilize ESPCN~\cite{shi2016real} in UPNet followed by one Conv layer. The output of RDN can be obtained by
\begin{align}
\begin{split}
\label{eq:I_SR}
I_{SR}=H_{RDN}\left ( I_{LR} \right ),
\end{split}
\end{align}  
where $H_{RDN}$ denotes the function of our RDN.
\begin{figure}[htpb]
\centering
\includegraphics[scale = 0.8]{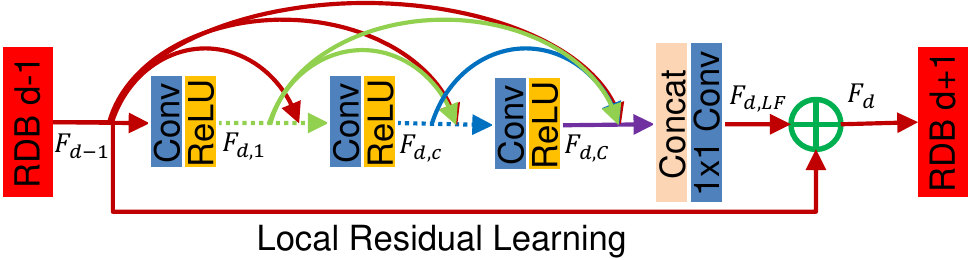}
\caption{Residual dense block (RDB) architecture.}
\label{fig:RDB} 
\vspace{-4mm} 
\end{figure}

\subsection{Residual Dense Block}
\label{subsec:RDB}
   
Now we present details about our proposed residual dense block (RDB) in Fig.~\ref{fig:RDB}. Our RDB contains dense connected layers, local feature fusion (LFF), and local residual learning, leading to a contiguous memory (CM) mechanism.     

\textbf{Contiguous memory} mechanism is realized by passing the state of preceding RDB to each layer of current RDB. Let $F_{d-1}$ and $F_{d}$ be the input and output of the $d$-th RDB respectively and both of them have G$_{0}$ feature-maps. The output of $c$-th Conv layer of $d$-th RDB can be formulated as
\begin{align}
\begin{split}
\label{eq:F_d_c}
F_{d,c}=\sigma \left ( W_{d,c}\left [ F_{d-1},F_{d,1},\cdots ,F_{d,c-1} \right ]  \right ),
\end{split}
\end{align}
where $\sigma$ denotes the ReLU~\cite{glorot2011deep} activation function. $W_{d,c}$ is the weights of the $c$-th Conv layer, where the bias term is omitted for simplicity. We assume $F_{d,c}$ consists of G (also known as growth rate~\cite{huang2017densely}) feature-maps. $\left [ F_{d-1},F_{d,1},\cdots ,F_{d,c-1} \right ]$ refers to the concatenation of the feature-maps produced by the $(d-1)$-th RDB, convolutional layers $1,\cdots ,\left ( c-1 \right )$ in the $d$-th RDB, resulting in G$_{0}$+$\left ( c-1 \right )\times $G feature-maps. The outputs of the preceding RDB and each layer have direct connections to all subsequent layers, which not only preserves the feed-forward nature, but also extracts local dense feature.

\textbf{Local feature fusion} is then applied to adaptively fuse the states from preceding RDB and the whole Conv layers in current RDB. As analyzed above, the feature-maps of the $\left ( d-1 \right )$-th RDB are introduced directly to the $d$-th RDB in a concatenation way, it is essential to reduce the feature number. On the other hand, inspired by~MemNet~\cite{tai2017memnet}, we introduce a $1\times 1$ convolutional layer to adaptively control the output information. We name this operation as local feature fusion (LFF) formulated as
\begin{align}
\begin{split}
\label{eq:F_d_LF}
F_{d,LF}=H_{LFF}^{d}\left ( \left [ F_{d-1},F_{d,1},\cdots ,F_{d,c},\cdots ,F_{d,C}\right ] \right ),
\end{split}
\end{align}
where $H_{LFF}^{d}$ denotes the function of the $1\times 1$ Conv layer in the $d$-th RDB. We also find that as the growth rate G becomes larger, very deep dense network without LFF would be hard to train.

\textbf{Local residual learning} is introduced in RDB to further improve the information flow, as there are several convolutional layers in one RDB. The final output of the $d$-th RDB can be obtained by
\begin{align}
\begin{split}
\label{eq:F_d_RDB}
F_{d}=F_{d-1}+F_{d,LF}.
\end{split}
\end{align}
It should be noted that LRL can also further improve the network representation ability, resulting better performance. We introduce more results about LRL in Section~\ref{sec:results}. Because of the dense connectivity and local residual learning, we refer to this block architecture as residual dense block (RDB). More differences between RDB and original dense block~\cite{huang2017densely} would be summarized in Section~\ref{sec:discussions}.

\subsection{Dense Feature Fusion}
\label{subsec:DFF}
After extracting local dense features with a set of RDBs, we further propose dense feature fusion (DFF) to exploit hierarchical features in a global way. Our DFF consists of global feature fusion (GFF) and global residual learning. 

\textbf{Global feature fusion} is proposed to extract the global feature $F_{GF}$ by fusing features from all the RDBs
\begin{align}
\begin{split}
\label{eq:GFF}
F_{GF}=H_{GFF}\left ( \left [ F_{1},\cdots ,F_{D} \right ] \right ),
\end{split}
\end{align}
where $\left [ F_{1},\cdots ,F_{D} \right ]$ refers to the concatenation of feature-maps produced by residual dense blocks $1,\cdots ,D$. $H_{GFF}$ is a composite function of $1\times 1$ and $3\times 3$ convolution. The $1\times 1$ convolutional layer is used to adaptively fuse a range of features with different levels. The following $3\times 3$ convolutional layer is introduced to further extract features for global residual learning, which has been demonstrated to be effective in~\cite{ledig2017photo}.

\textbf{Global residual learning} is then utilized to obtain the feature-maps before conducting up-scaling by 
\begin{align}
\begin{split}
\label{eq:DFF_GF}
F_{DF}=F_{-1}+F_{GF},
\end{split}
\end{align}
where $F_{-1}$ denotes the shallow feature-maps. All the other layers before global feature fusion are fully utilized with our proposed residual dense blocks (RDBs). RDBs produce multi-level local dense features, which are further adaptively fused to form $F_{GF}$. After global residual learning, we obtain dense feature $F_{DF}$.

It should be noted that Tai~et al.~\cite{tai2017memnet} utilized long-term dense connections in MemNet to recover more high frequency information. However, in the memory block~\cite{tai2017memnet}, the preceding layers don't have direct access to all the subsequent layers. The local feature information are not fully used, limiting the ability of long-term connections. In addition, MemNet extracts features in the HR space, increasing computational complexity. While, inspired by~\cite{dong2016accelerating,shi2016real,lai2017deep,lim2017enhanced}, we extract local and global features in the LR space. More differences between our residual dense network and MemNet would be shown in Section~\ref{sec:discussions}. We would also demonstrate the effectiveness of global feature fusion in Section~\ref{sec:results}.

\subsection{Implementation Details}
\label{sec:implement}
In our proposed RDN, we set $3\times 3$ as the size of all convolutional layers except that in local and global feature fusion, whose kernel size is $1\times 1$. For convolutional layer with kernel size $3\times 3$, we pad zeros to each side of the input to keep size fixed. Shallow feature extraction layers, local and global feature fusion layers have G$_{0}$=64 filters. Other layers in each RDB has G filters and are followed by ReLU~\cite{glorot2011deep}. Following~\cite{lim2017enhanced}, we use ESPCNN~\cite{shi2016real} to upscale the coarse resolution features to fine ones for the UPNet. The final Conv layer has $3$ output channels, as we output color HR images. However, the network can also process gray images. 

\section{Discussions}
\label{sec:discussions}

\textbf{Difference to DenseNet.} Inspired from DenseNet~\cite{huang2017densely}, we adopt the local dense connections into our proposed residual dense block (RDB). In general, DenseNet is widely used in high-level computer vision tasks (e.g., object recognition). While RDN is designed for image SR. Moreover, we remove batch nomalization (BN) layers, which consume the same amount of GPU memory as convolutional layers, increase computational complexity, and hinder performance of the network. We also remove the pooling layers, which could discard some pixel-level information. Furthermore, transition layers are placed into two adjacent dense blocks in DenseNet. While in RDN, we combine dense connected layers with local feature fusion (LFF) by using local residual learning, which would be demonstrated to be effective in Section~\ref{sec:results}. As a result, the output of the $(d-1)$-th RDB has direct connections to each layer in the $d$-th RDB and also contributes to the input of $(d+1)$-th RDB. Last not the least, we adopt global feature fusion to fully use hierarchical features, which are neglected in DenseNet.  

\textbf{Difference to SRDenseNet.} There are three main differences between SRDenseNet~\cite{tong2017image} and our RDN. The first one is the design of basic building block. SRDenseNet introduces the basic dense block from DenseNet~\cite{huang2017densely}. Our residual dense block (RDB) improves it in three ways: (1). We introduce contiguous memory (CM) mechanism, which allows the state of preceding RDB have direct access to each layer of the current RDB. (2). Our RDB allow larger growth rate by using local feature fusion (LFF), which stabilizes the training of wide network. (3). Local residual learning (LRL) is utilized in RDB to further encourage the flow of information and gradient. The second one is there is no dense connections among RDB. Instead we use global feature fusion (GFF) and global residual learning to extract global features, because our RDBs with contiguous memory have fully extracted features locally. As shown in Sections~\ref{subsec:study_DCG} and~\ref{subsec:ablation}, all of these components increase the performance significantly. The third one is SRDenseNet uses $L_2$ loss function. Whereas we utilize $L_1$ loss function, which has been demonstrated to be more powerful for performance and convergence~\cite{lim2017enhanced}. As a result, our proposed RDN achieves better performance than that of SRDenseNet. 

\textbf{Difference to MemNet.} In addition to the different choice of loss function ($L_2$ in MemNet~\cite{tai2017memnet}),  we mainly summarize another three differences bwtween MemNet and our RDN. First, MemNet needs to upsample the original LR image to the desired size using Bicubic interpolation. This procedure results in feature extraction and reconstruction in HR space. While, RDN extracts hierarchical features from the original LR image, reducing computational complexity significantly and improving the performance. Second, the memory block in MemNet contains recursive and gate units. Most layers within one recursive unit don't receive the information from their preceding layers or memory block. While, in our proposed RDN, the output of RDB has direct access to each layer of the next RDB. Also the information of each convolutional layer flow into all the subsequent layers within one RDB. Furthermore, local residual learning in RDB improves the flow of information and gradients and performance, which is demonstrated in Section~\ref{sec:results}. Third, as analyzed above, current memory block doesn't fully make use of the information of the output of the preceding block and its layers. Even though MemNet adopts densely connections among memory blocks in the HR space, MemNet fails to fully extract hierarchical features from the original LR inputs. While, after extracting local dense features with RDBs, our RDN further fuses the hierarchical features from the whole preceding layers in a global way in the LR space. 

\section{Experimental Results}
\label{sec:results}
\vspace{-2mm}
The source code of the proposed method can be downloaded at \href{https://github.com/yulunzhang/RDN}{https://github.com/yulunzhang/RDN}.

\subsection{Settings}
\label{subsec:settings}
\vspace{-2mm}
\textbf{Datasets and Metrics.}
Recently, Timofte et al. have released a high-quality (2K resolution) dataset DIV2K for image restoration applications~\cite{timofte2017ntire}. DIV2K consists of 800 training images, 100 validation images, and 100 test images. We train all of our models with 800 training images and use 5 validation images in the training process. For testing, we use five standard benchmark datasets: Set5~\cite{bevilacqua2012low}, Set14~\cite{zeyde2012single}, B100~\cite{martin2001database}, Urban100~\cite{huang2015single}, and Manga109~\cite{matsui2017sketch}. The SR results are evaluated with PSNR and SSIM~\cite{wang2004image} on Y channel (\textit{i.e.}, luminance) of transformed YCbCr space.

\textbf{Degradation Models.} In order to fully demonstrate the effectiveness of our proposed RDN, we use three degradation models to simulate LR images. The first one is bicubic downsampling by adopting the Matlab function \textit{imresize} with the option \textit{bicubic} (denote as \textbf{BI} for short). We use \textbf{BI} model to simulate LR images with scaling factor $\times2$, $\times3$, and $\times4$. Similar to~\cite{zhang2017learning}, the second one is to blur HR image by Gaussian kernel of size $7\times7$ with standard deviation 1.6. The blurred image is then downsampled with scaling factor $\times3$ (denote as \textbf{BD} for short). We further produce LR image in a more challenging way. We first bicubic downsample HR image with scaling factor $\times3$ and then add Gaussian noise with noise level 30 (denote as \textbf{DN} for short). 

\textbf{Training Setting.}
Following settings of~\cite{lim2017enhanced}, in each training batch, we randomly extract 16 LR RGB patches with the size of $32\times32$ as inputs. We randomly augment the patches by flipping horizontally or vertically and rotating 90$^{\circ}$. 1,000 iterations of back-propagation constitute an epoch. We implement our RDN with the Torch7 framework and update it with Adam optimizer~\cite{kingma2014adam}. The learning rate is initialized to 10$^{-4}$ for all layers and decreases half for every 200 epochs. Training a RDN roughly takes 1 day with a Titan Xp GPU for 200 epochs.
\vspace{-4mm}
\begin{figure}[htbp]
\scriptsize
\centering
\centerline{
\subfigure[]{
\label{fig:investeD}
\includegraphics[width = 27mm]{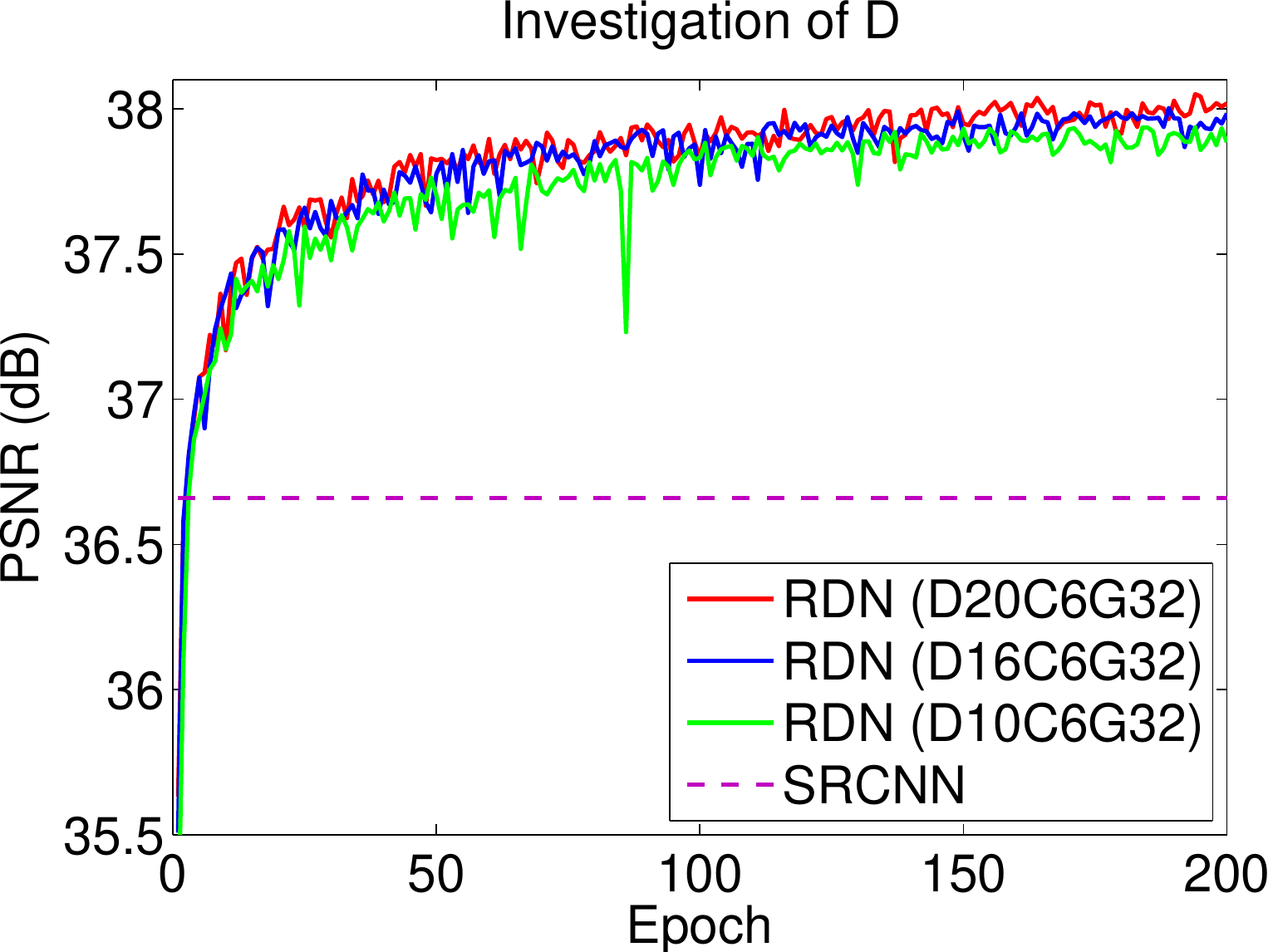}}
\subfigure[]{
\label{fig:investeC}
\includegraphics[width = 27mm]{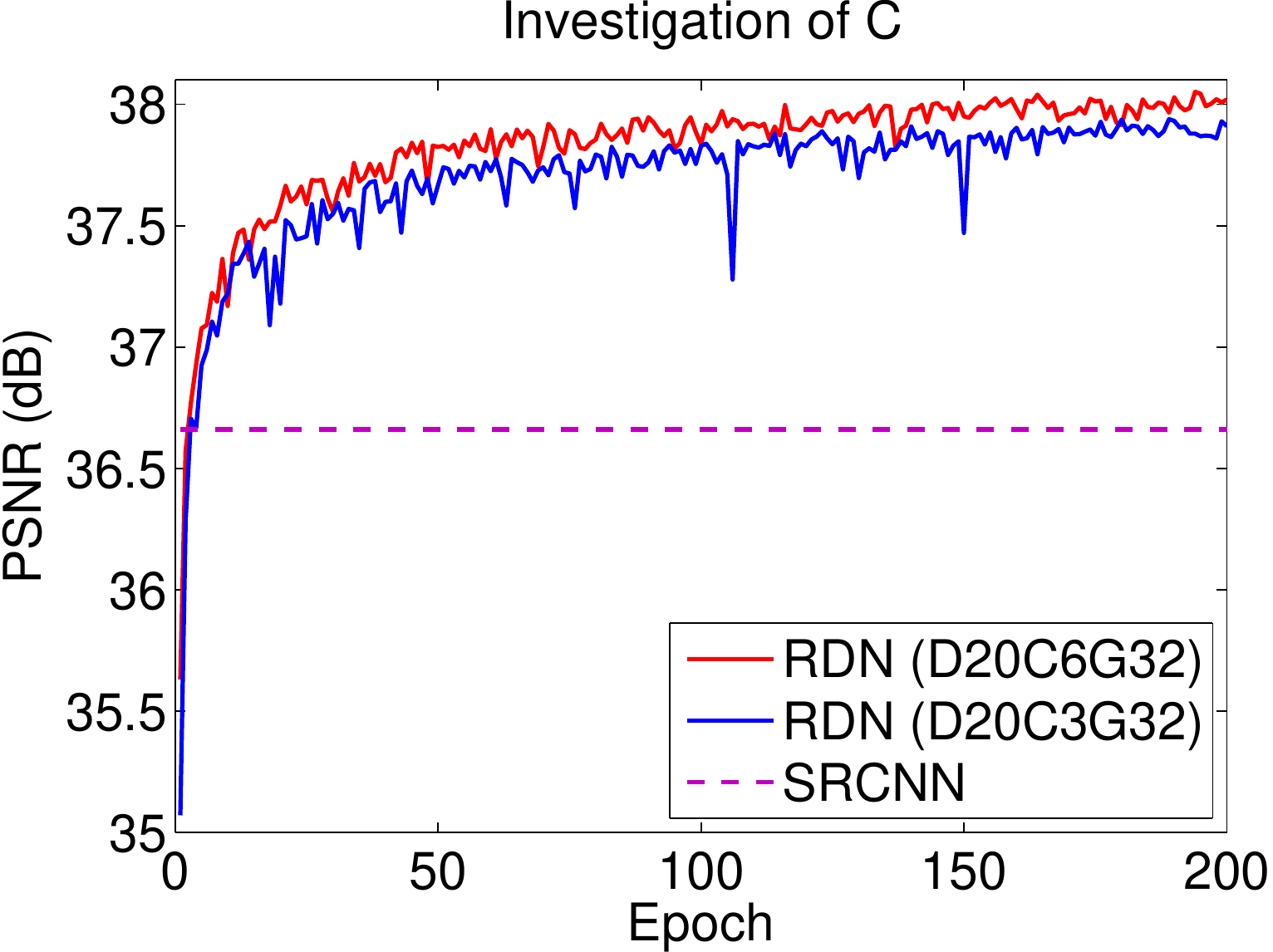}}
\subfigure[]{
\label{fig:investeG}
\includegraphics[width = 27mm]{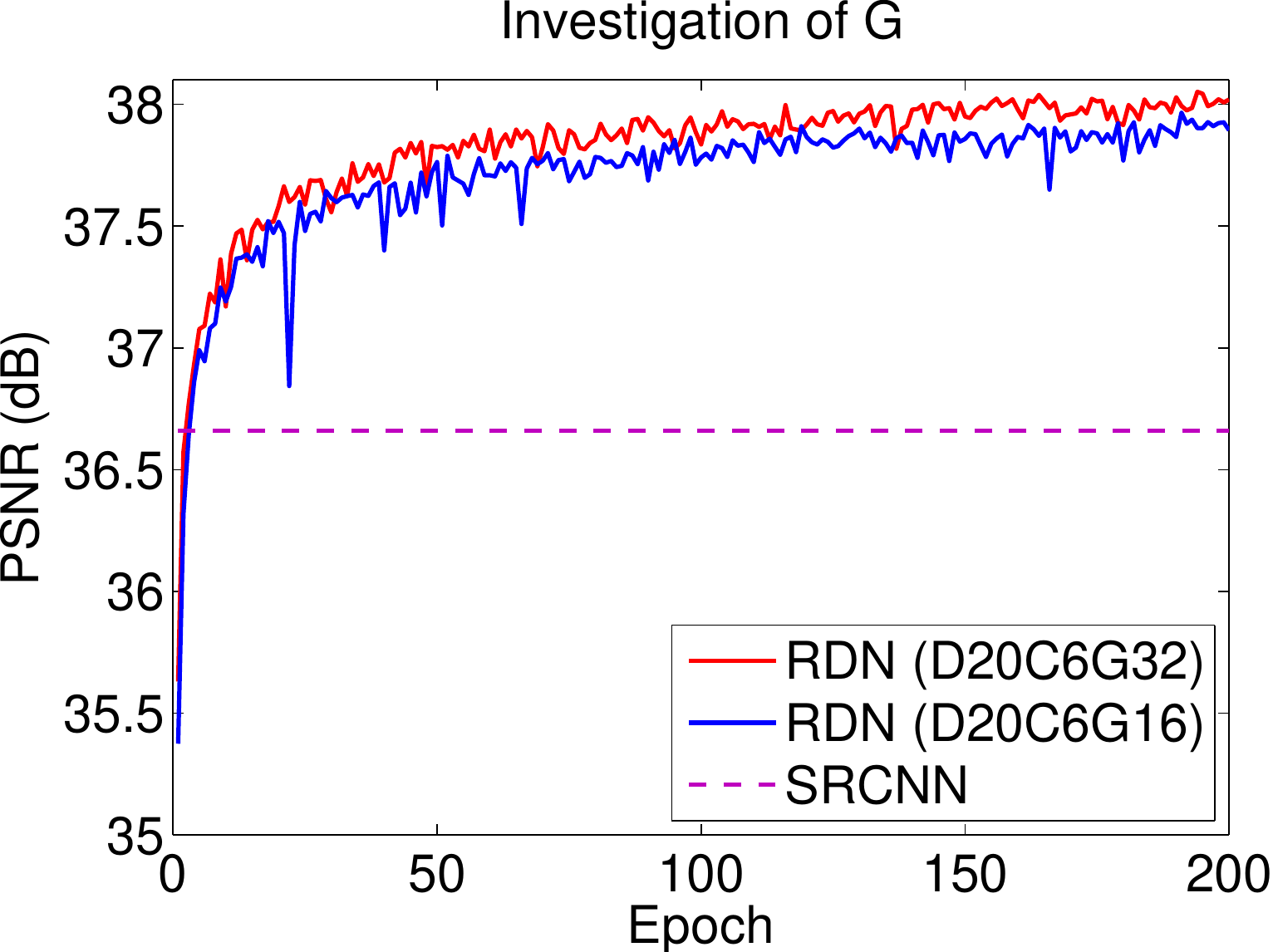}}
}
\vspace{-1mm}
\caption{Convergence analysis of RDN with different values of D, C, and G.}  
\label{fig:study_D_C_G}
\vspace{-4mm}
\end{figure}

\subsection{Study of D, C, and G.}
\label{subsec:study_DCG}
In this subsection, we investigate the basic network parameters: the number of RDB (denote as D for short), the number of Conv layers per RDB (denote as C for short), and the growth rate (denote as G for short). We use the performance of SRCNN~\cite{dong2016image} as a reference. As shown in Figs.~\ref{fig:investeD} and~\ref{fig:investeC}, larger D or C would lead to higher performance. This is mainly because the network becomes deeper with larger D or C. As our proposed LFF allows larger G, we also observe larger G (see Fig.~\ref{fig:investeG}) contributes to better performance. On the other hand, RND with smaller D, C, or G would suffer some performance drop in the training, but RDN would still outperform SRCNN~\cite{dong2016image}. More important, our RDN allows deeper and wider network, from which more hierarchical features are extracted for higher performance.       

\begin{table}[tbp]
\scriptsize
\centering
\begin{center}

\begin{tabular*}{82.4mm}{@{\extracolsep{-0.75mm}}|c|c|c|c|c|c|c|c|c|}
\hline
 & \multicolumn{8}{c|}{Different combinations of CM, LRL, and GFF} 
\\ 
\hline  
\hline
CM  & \XSolid & \Checkmark & \XSolid & \XSolid & \Checkmark & \Checkmark & \XSolid & \Checkmark
\\
LRL & \XSolid & \XSolid   & \Checkmark & \XSolid & \Checkmark & \XSolid & \Checkmark & \Checkmark
\\
GFF & \XSolid & \XSolid   & \XSolid & \Checkmark & \XSolid & \Checkmark & \Checkmark & \Checkmark
\\
\hline
\hline
PSNR & 34.87 & 37.89 & 37.92 & 37.78 & 37.99 & 37.98 & 37.97 & 38.06 
\\
\hline
\end{tabular*}
\end{center}
\vspace{-3mm}
\caption{Ablation investigation of contiguous memory (CM), local residual learning (LRL), and global feature fusion (GFF). We observe the best performance (PSNR) on Set5 with scaling factor $\times2$ in 200 epochs.}
\label{tab:results_ablation}
\vspace{-6mm}
\end{table}

\subsection{Ablation Investigation}
\label{subsec:ablation}
Table~\ref{tab:results_ablation} shows the ablation investigation on the effects of contiguous memory (CM), local residual learning (LRL), and global feature fusion (GFF). The eight networks have the same RDB number (D = 20), Conv number (C = 6) per RDB, and growth rate (G = 32). We find that local feature fusion (LFF) is needed to train these networks properly, so LFF isn't removed by default. The baseline (denote as RDN\_CM0LRL0GFF0) is obtained without CM, LRL, or GFF and performs very poorly (PSNR = 34.87 dB). This is caused by the difficulty of training~\cite{dong2016image} and also demonstrates that stacking many basic dense blocks~\cite{huang2017densely} in a very deep network would not result in better performance.      

We then add one of CM, LRL, or GFF to the baseline, resulting in RDN\_CM1LRL0GFF0, RDN\_CM0LRL1GFF0, and RDN\_CM0LRL0GFF1 respectively (from 2$^{nd}$ to 4$^{th}$ combination in Table~\ref{tab:results_ablation}). We can validate that each component can efficiently improve the performance of the baseline. This is mainly because each component contributes to the flow of information and gradient. 

We further add two components to the baseline, resulting in RDN\_CM1LRL1GFF0, RDN\_CM1LRL0GFF1, and RDN\_CM0LRL1GFF1 respectively (from 5$^{th}$ to 7$^{th}$ combination in Table~\ref{tab:results_ablation}). It can be seen that two components would perform better than only one component. Similar phenomenon can be seen when we use these three components simultaneously (denote as RDN\_CM1LRL1GFF1). RDN using three components performs the best.

\begin{figure}[t]
\scriptsize
\centering
\centerline{
\includegraphics[scale =0.40]{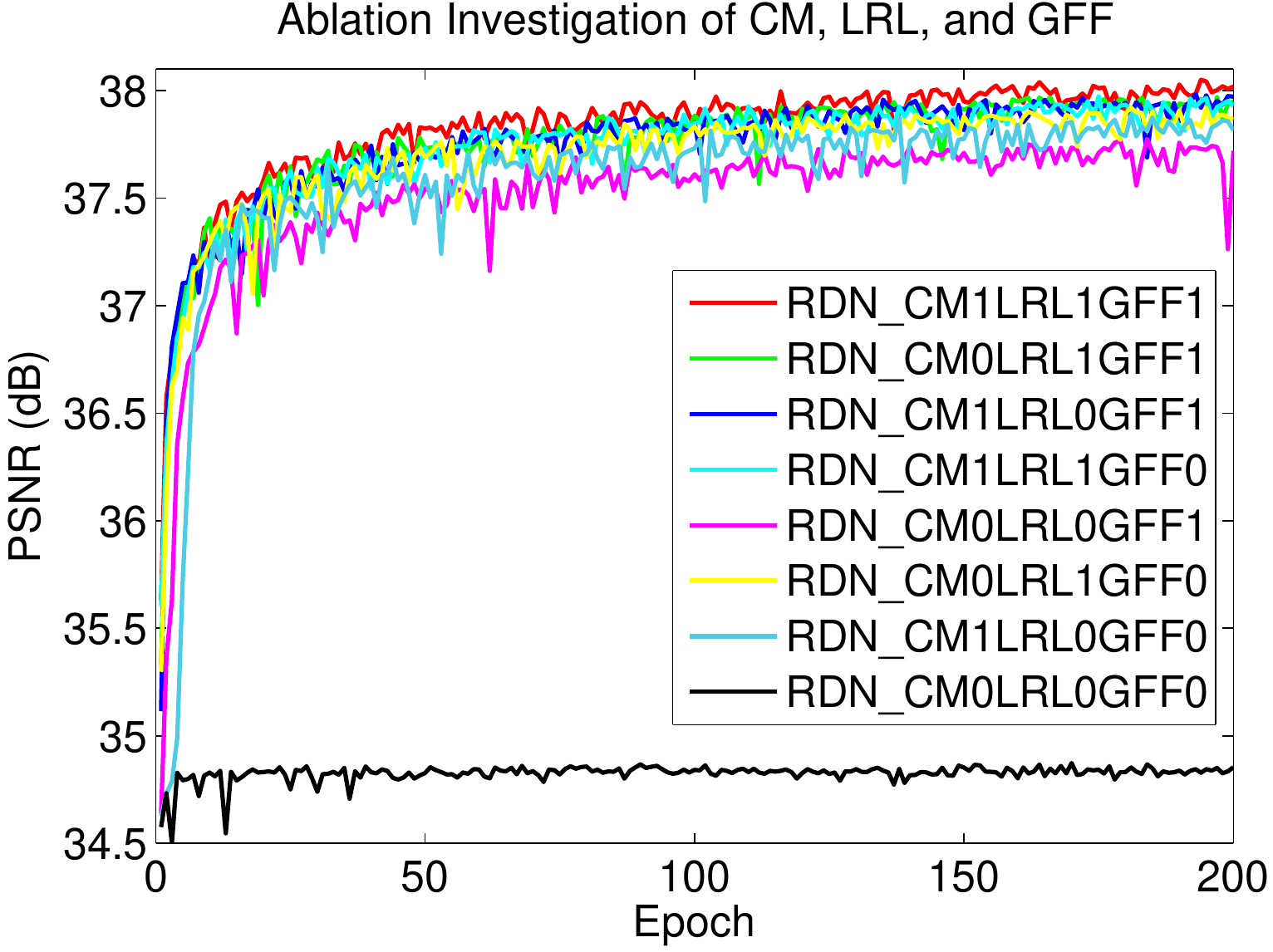}}
\caption{Convergence analysis on CM, LRL, and GFF. The curves for each combination are based on the PSNR on Set5 with scaling factor~$\times2$ in 200 epochs.}  
\label{fig:study_CM_LRL_GFF}
\vspace{-5mm}
\end{figure}

We also visualize the convergence process of these eight combinations in Fig.~\ref{fig:study_CM_LRL_GFF}. The convergence curves are consistent with the analyses above and show that CM, LRL, and GFF can further stabilize the training process without obvious performance drop. These quantitative and visual analyses demonstrate the effectiveness and benefits of our proposed CM, LRL, and GFF.

\begin{table*}[htpb]
\scriptsize
\centering
\begin{center}

\begin{tabular*}{169.85mm}{@{\extracolsep{-0.928mm}}|c|c|c|c|c|c|c|c|c|c|c|c|c|c|c|c|c|}
\hline
\multirow{2}{*}{Dataset} & \multirow{2}{*}{Scale} &  \multirow{2}{*}{Bicubic} & SRCNN &  LapSRN &  DRRN &  SRDenseNet &  MemNet & MDSR & RDN & RDN+ 
\\
&  &  & \cite{dong2016image} & \cite{lai2017deep} & \cite{tai2017image} & \cite{tong2017image} & \cite{tai2017memnet} & \cite{lim2017enhanced} & (ours) & (ours)    
\\
\hline
\hline
\multirow{3}{*}{Set5}
& $\times2$ 
 & 33.66/0.9299
  & 36.66/0.9542
   & 37.52/0.9591
    & 37.74/0.9591
     & -/- 
      & 37.78/0.9597
       & 38.11/0.9602
        & 38.24/0.9614
         & \textbf{38.30}/\textbf{0.9616}
                        
\\
& $\times3$ 
& 30.39/0.8682
 & 32.75/0.9090
  & 33.82/0.9227
   & 34.03/0.9244
    & -/-
     & 34.09/0.9248
      & 34.66/0.9280
       & 34.71/0.9296
        & \textbf{34.78}/\textbf{0.9300}
                            
\\
& $\times4$ 
& 28.42/0.8104
 & 30.48/0.8628
  & 31.54/0.8855
   & 31.68/0.8888
    & 32.02/0.8934
     & 31.74/0.8893
      & 32.50/0.8973
       & 32.47/0.8990
        & \textbf{32.61}/\textbf{0.9003}
                              
\\
\hline 
\hline
\multirow{3}{*}{Set14}
& $\times2$ 
& 30.24/0.8688
 & 32.45/0.9067
  & 33.08/0.9130
   & 33.23/0.9136
    & -/-
     & 33.28/0.9142
      & 33.85/0.9198
       & 34.01/0.9212
        & \textbf{34.10}/\textbf{0.9218}
                      
\\
& $\times3$ 
& 27.55/0.7742
 & 29.30/0.8215
  & 29.79/0.8320
   & 29.96/0.8349
    & -/-
     & 30.00/0.8350
      & 30.44/0.8452
       & 30.57/0.8468
        & \textbf{30.67}/\textbf{0.8482}
                            
\\
& $\times4$ 
& 26.00/0.7027
 & 27.50/0.7513
  & 28.19/0.7720
   & 28.21/0.7721
    & 28.50/0.7782
     & 28.26/0.7723
      & 28.72/0.7857
       & 28.81/0.7871
        & \textbf{28.92}/\textbf{0.7893}
                              
\\
\hline
\hline
\multirow{3}{*}{B100}
& $\times2$ 
& 29.56/0.8431
 & 31.36/0.8879
  & 31.80/0.8950
   & 32.05/0.8973
    & -/-
     & 32.08/0.8978
      & 32.29/0.9007
       & 32.34/0.9017
        & \textbf{32.40}/\textbf{0.9022}
                      
\\
& $\times3$ 
& 27.21/0.7385
 & 28.41/0.7863
  & 28.82/0.7973
   & 28.95/0.8004
    & -/-
     & 28.96/0.8001
      & 29.25/0.8091
       & 29.26/0.8093
        & \textbf{29.33}/\textbf{0.8105}
                            
\\
& $\times4$ 
& 25.96/0.6675
 & 26.90/0.7101
  & 27.32/0.7280
   & 27.38/0.7284
    & 27.53/0.7337
     & 27.40/0.7281
      & 27.72/0.7418
       & 27.72/0.7419
        & \textbf{27.80}/\textbf{0.7434}
                              
\\
\hline
\hline
\multirow{3}{*}{Urban100}
& $\times2$ 
& 26.88/0.8403
 & 29.50/0.8946
  & 30.41/0.9101
   & 31.23/0.9188
    & -/-
     & 31.31/0.9195
      & 32.84/0.9347
       & 32.89/0.9353
        & \textbf{33.09}/\textbf{0.9368}
                      
\\
& $\times3$ 
& 24.46/0.7349
 & 26.24/0.7989
  & 27.07/0.8272
   & 27.53/0.8378
    & -/-
     & 27.56/0.8376
      & 28.79/0.8655
       & 28.80/0.8653
        & \textbf{29.00}/\textbf{0.8683}
                            
\\
& $\times4$ 
& 23.14/0.6577
 & 24.52/0.7221
  & 25.21/0.7553
   & 25.44/0.7638
    & 26.05/0.7819
     & 25.50/0.7630
      & 26.67/0.8041
       & 26.61/0.8028
        & \textbf{26.82}/\textbf{0.8069}
                              
\\
\hline
\hline
\multirow{3}{*}{Manga109}
& $\times2$ 
& 30.80/0.9339
 & 35.60/0.9663
  & 37.27/0.9740
   & 37.60/0.9736
    & -/-
     & 37.72/0.9740
      & 38.96/0.9769
       & 39.18/0.9780
        & \textbf{39.38}/\textbf{0.9784}
                      
\\
& $\times3$ 
& 26.95/0.8556
 & 30.48/0.9117
  & 32.19/0.9334
   & 32.42/0.9359
    & -/-
     & 32.51/0.9369
      & 34.17/0.9473
       & 34.13/0.9484
        & \textbf{34.43}/\textbf{0.9498}
                            
\\
& $\times4$ 
& 24.89/0.7866
 & 27.58/0.8555
  & 29.09/0.8893
   & 29.18/0.8914
    & -/-
     & 29.42/0.8942
      & 31.11/0.9148
       & 31.00/0.9151
        & \textbf{31.39}/\textbf{0.9184}
                              
\\
\hline          
\end{tabular*}
\end{center}
\vspace{-2mm}
\caption{Benchmark results with \textbf{BI} degradation model. Average PSNR/SSIM values for scaling factor $\times2$, $\times3$, and $\times4$.}
\label{tab:results_BI_5sets}
\vspace{-3mm}
\end{table*}
\begin{figure*}[htbp]
\centering
\includegraphics[width = 170mm]{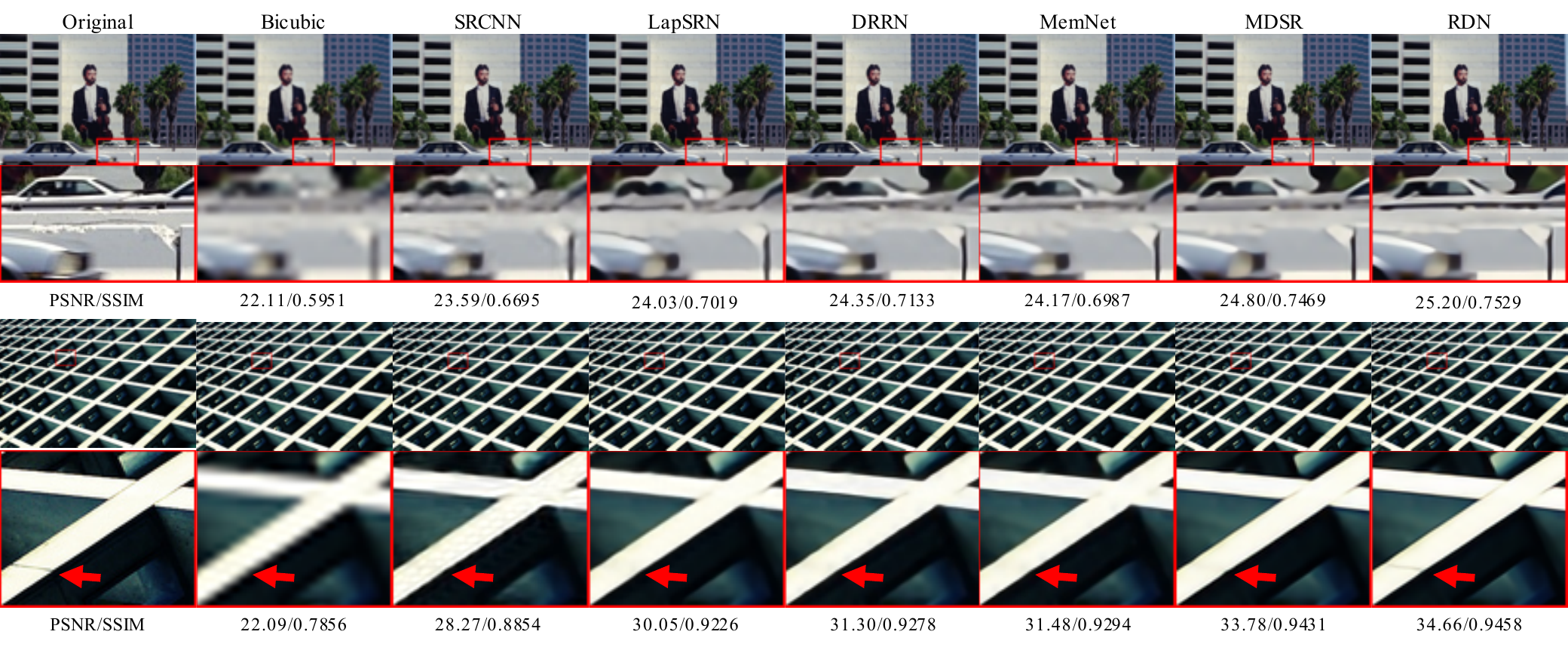}
\caption{Visual results with \textbf{BI} model ($\times4$). The SR results are for image ``119082" from B100 and ``img\_043'' from Urban100 respectively.}
\label{fig:visual_BI}  
\vspace{-5mm}
\end{figure*}
\subsection{Results with BI Degradation Model}
\label{subsec:BI-degradation}
Simulating LR image with BI degradation model is widely used in image SR settings. For BI degradation model, we compare our RDN with 6 state-of-the-art image SR methods: SRCNN~\cite{dong2016image}, LapSRN~\cite{lai2017deep}, DRRN~\cite{tai2017image}, SRDenseNet~\cite{tong2017image}, MemNet~\cite{tai2017memnet}, and MDSR~\cite{lim2017enhanced}. Similar to~\cite{timofte2016seven,lim2017enhanced}, we also adopt self-ensemble strategy~\cite{lim2017enhanced} to further improve our RDN and denote the self-ensembled RDN as RDN+. As analyzed above, a deeper and wider RDN would lead to a better performance. On the other hand, as most methods for comparison only use about 64 filters per Conv layer, we report results of RDN by using D = 16, C = 8, and G = 64 for fair comparison. EDSR~\cite{lim2017enhanced} is skipped here, because it uses far more filters (\textit{i.e.}, 256) per Conv layer, leading to a very wide network with high number of parameters. However, our RDN would also achieve comparable or even better results than those by EDSR~\cite{lim2017enhanced}.

Table~\ref{tab:results_BI_5sets} shows quantitative comparisons for $\times2$, $\times3$, and $\times4$ SR. Results of SRDenseNet~\cite{tong2017image} are cited from their paper. When compared with persistent CNN models ( SRDenseNet~\cite{tong2017image} and MemNet~\cite{tai2017memnet}), our RDN performs the best on all datasets with all scaling factors. This indicates the better effectiveness of our residual dense block (RDB) over dense block in SRDensenet~\cite{tong2017image} and memory block in MemNet~\cite{tai2017memnet}. When compared with the remaining models, our RDN also achieves the best average results on most datasets. Specifically, for the scaling factor $\times2$, our RDN performs the best on all datasets. When the scaling factor becomes larger (e.g., $\times3$ and $\times4$), RDN would not hold the similar advantage over MDSR~\cite{lim2017enhanced}. There are mainly three reasons for this case. First, MDSR is deeper (160 v.s. 128), having about 160 layers to extract features in LR space. Second, MDSR utilizes multi-scale inputs as VDSR does~\cite{kim2016accurate}. Third, MDSR uses larger input patch size (65 v.s. 32) for training. As most images in Urban100 contain self-similar structures, larger input patch size for training allows a very deep network to grasp more information by using large receptive field better. As we mainly focus on the effectiveness of our RDN and fair comparison, we don't use deeper network, multi-scale information, or larger input patch size. Moreover, our RDN+ can achieve further improvement with self-ensemble~\cite{lim2017enhanced}.

In Fig.~\ref{fig:visual_BI}, we show visual comparisons on scale $\times4$. For image ``119082'', we observe that most of compared methods would produce noticeable artifacts and produce blurred edges. In contrast, our RDN can recover sharper and clearer edges, more faithful to the ground truth. For the tiny line (pointed by the {\color{red}red arrow}) in image ``'img\_043', all the compared methods fail to recover it. While, our RDN can recover it obviously. This is mainly because RDN uses hierarchical features through dense feature fusion.   

\begin{table*}[htbp]
\scriptsize
\centering
\begin{center}

\begin{tabular*}{170.9mm}{@{\extracolsep{-0.928mm}}|c|c|c|c|c|c|c|c|c|c|c|c|c|c|c|c|c|}
\hline
\multirow{2}{*}{Dataset} & \multirow{2}{*}{Model} &  \multirow{2}{*}{Bicubic} & SPMSR & SRCNN &  FSRCNN &  VDSR &  IRCNN\_G &  IRCNN\_C & RDN & RDN+ 
\\
&  &  & \cite{peleg2014statistical} & \cite{dong2016image} & \cite{dong2016accelerating} & \cite{kim2016accurate} & \cite{zhang2017learning} & \cite{zhang2017learning} & (ours) & (ours)    
\\
\hline
\hline
\multirow{2}{*}{Set5}
& \textbf{BD} 
& 28.78/0.8308
 & 32.21/0.9001
  & 32.05/0.8944
   & 26.23/0.8124
    & 33.25/0.9150
     & 33.38/0.9182
      & 33.17/0.9157
       & 34.58/0.9280
        & \textbf{34.70}/\textbf{0.9289}
                            
\\
& \textbf{DN} 
& 24.01/0.5369
 & -/-
  & 25.01/0.6950
   & 24.18/0.6932
    & 25.20/0.7183
     & 25.70/0.7379
      & 27.48/0.7925
       & 28.47/0.8151
        & \textbf{28.55}/\textbf{0.8173}
                              
\\
\hline 
\hline
\multirow{2}{*}{Set14}
& \textbf{BD} 
& 26.38/0.7271
 & 28.89/0.8105
  & 28.80/0.8074
   & 24.44/0.7106
    & 29.46/0.8244
     & 29.63/0.8281
      & 29.55/0.8271
       & 30.53/0.8447
        & \textbf{30.64}/\textbf{0.8463}
                            
\\
& \textbf{DN} 
& 22.87/0.4724
 & -/-
  & 23.78/0.5898
   & 23.02/0.5856
    & 24.00/0.6112
     & 24.45/0.6305
      & 25.92/0.6932
       & 26.60/0.7101
        & \textbf{26.67}/\textbf{0.7117}

\\
\hline
\hline
\multirow{2}{*}{B100}
& \textbf{BD} 
& 26.33/0.6918
 & 28.13/0.7740
  & 28.13/0.7736
   & 24.86/0.6832
    & 28.57/0.7893
     & 28.65/0.7922
      & 28.49/0.7886
       & 29.23/0.8079
        & \textbf{29.30}/\textbf{0.8093}
                            
\\
& \textbf{DN}
& 22.92/0.4449
 & -/-
  & 23.76/0.5538
   & 23.41/0.5556
    & 24.00/0.5749
     & 24.28/0.5900
      & 25.55/0.6481
       & 25.93/0.6573
        & \textbf{25.97}/\textbf{0.6587}

\\
\hline
\hline
\multirow{2}{*}{Urban100}
& \textbf{BD} 
& 23.52/0.6862
 & 25.84/0.7856
  & 25.70/0.7770
   & 22.04/0.6745
    & 26.61/0.8136
     & 26.77/0.8154
      & 26.47/0.8081
       & 28.46/0.8582
        & \textbf{28.67}/\textbf{0.8612}
                            
\\
& \textbf{DN} 
& 21.63/0.4687
 & -/-
  & 21.90/0.5737
   & 21.15/0.5682
    & 22.22/0.6096
     & 22.90/0.6429
      & 23.93/0.6950
       & 24.92/0.7364
        & \textbf{25.05}/\textbf{0.7399}
                              
\\
\hline
\hline
\multirow{2}{*}{Manga109}
& \textbf{BD} 
& 25.46/0.8149
 & 29.64/0.9003
  & 29.47/0.8924
   & 23.04/0.7927
    & 31.06/0.9234
     & 31.15/0.9245
      & 31.13/0.9236
       & 33.97/0.9465
        & \textbf{34.34}/\textbf{0.9483}
                            
\\
& \textbf{DN} 
& 23.01/0.5381
 & -/-
  & 23.75/0.7148
   & 22.39/0.7111
    & 24.20/0.7525
     & 24.88/0.7765
      & 26.07/0.8253
       & 28.00/0.8591
        & \textbf{28.18}/\textbf{0.8621}
                              
\\
\hline      
    
\end{tabular*}
\end{center}
\vspace{-3mm}
\caption{Benchmark results with \textbf{BD} and \textbf{DN} degradation models. Average PSNR/SSIM values for scaling factor $\times3$.} 
\label{tab:results_BD_DN_5sets}
\vspace{-5mm}
\end{table*}

\subsection{Results with BD and DN Degradation Models}
\label{subsec:BD-DN-degradation}
\begin{figure}[htbp]
\centering
\includegraphics[width = 84mm]{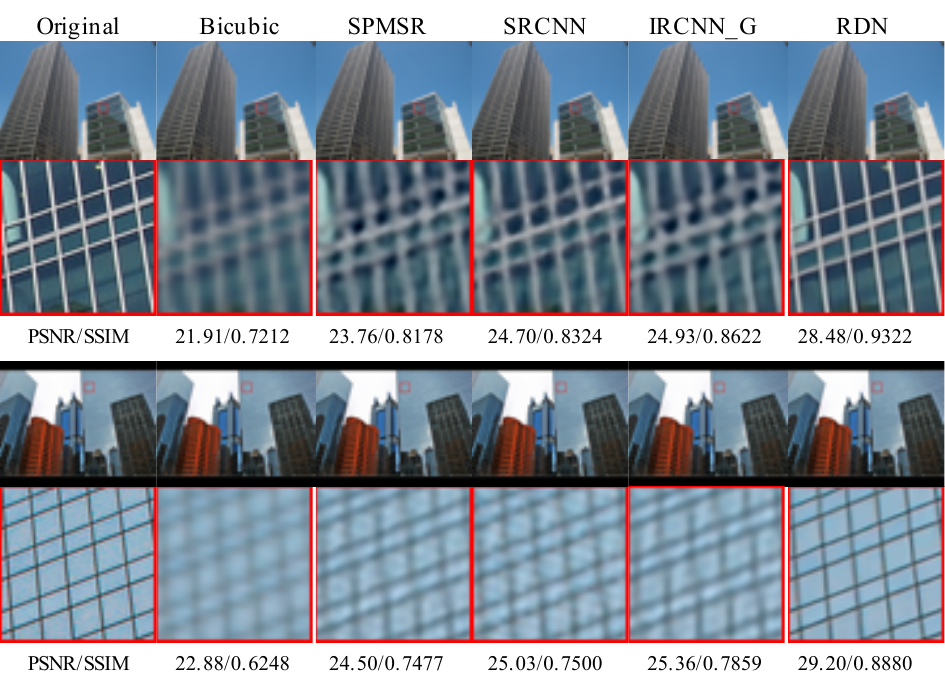}
\caption{Visual results using \textbf{BD} degradation model with scaling factor $\times3$. The SR results are for image ``img\_096" from Urban100 and ``img\_099'' from Urban100 respectively.}
\label{fig:visual_BD}
\vspace{-3mm}  
\end{figure}

\begin{figure}[htbp]
\centering
\includegraphics[width = 84mm]{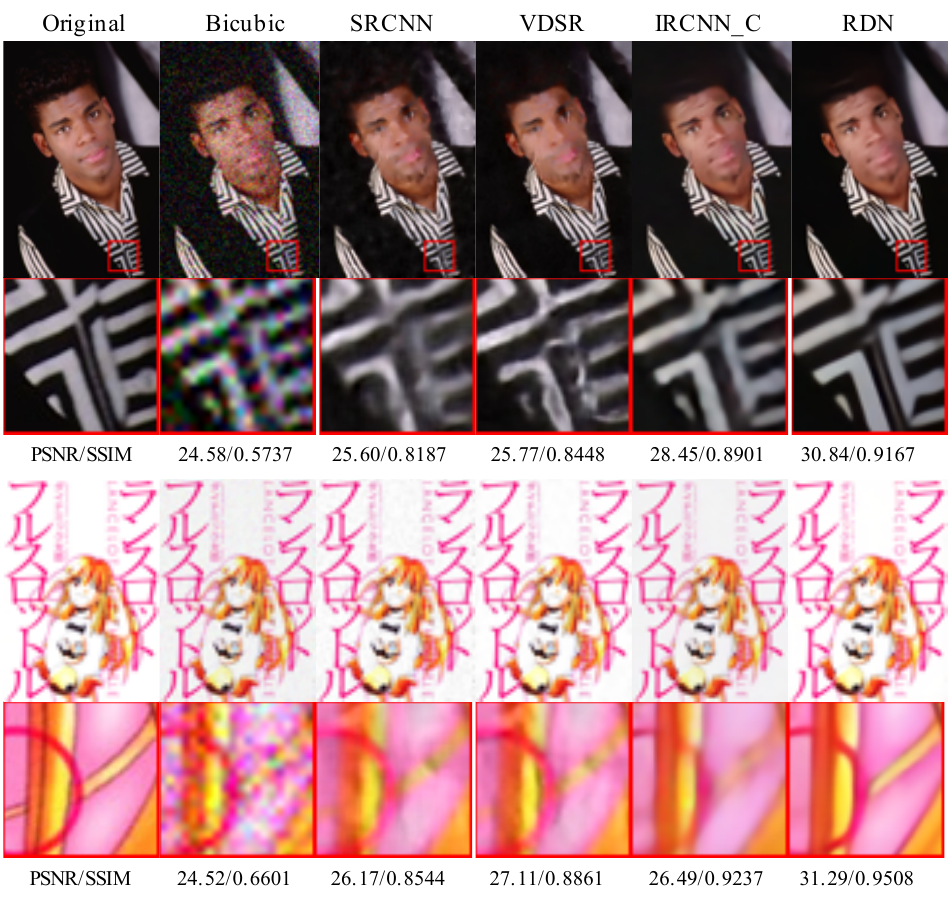}
\caption{Visual results using \textbf{DN} degradation model with scaling factor $\times3$. The SR results are for image ``302008" from B100 and ``LancelotFullThrottle'' from Manga109 respectively.}
\label{fig:visual_DN}  
\vspace{-5mm}
\end{figure}

Following~\cite{zhang2017learning}, we also show the SR results with BD degradation model and further introduce DN degradation model. Our RDN is compared with SPMSR~\cite{peleg2014statistical}, SRCNN~\cite{dong2016image}, FSRCNN~\cite{dong2016accelerating}, VDSR~\cite{kim2016accurate}, IRCNN\_G~\cite{zhang2017learning}, and IRCNN\_C~\cite{zhang2017learning}. We re-train SRCNN, FSRCNN, and VDSR for each degradation model. Table~\ref{tab:results_BD_DN_5sets} shows the average PSNR and SSIM results on Set5, Set14, B100, Urban100, and Manga109 with scaling factor $\times3$. Our RDN and RDN+ perform the best on all the datasets with BD and DN degradation models. The performance gains over other state-of-the-art methods are consistent with the visual results in Figs.~\ref{fig:visual_BD} and~\ref{fig:visual_DN}. 

For \textbf{BD} degradation model (Fig.~\ref{fig:visual_BD}), the methods using interpolated LR image as input would produce noticeable artifacts and be unable to remove the blurring artifacts. In contrast, our RDN suppresses the blurring artifacts and recovers sharper edges. This comparison indicates that extracting hierarchical features from the original LR image would alleviate the blurring artifacts. It also demonstrates the strong ability of RDN for \textbf{BD} degradation model.   

For \textbf{DN} degradation model (Fig.~\ref{fig:visual_DN}), where the LR image is corrupted by noise and loses some details. We observe that the noised details are hard to recovered by other methods~\cite{dong2016image,kim2016accurate,zhang2017learning}. However, our RDN can not only handle the noise efficiently, but also recover more details. This comparison indicates that RDN is applicable for jointly image denoising and SR. These results with \textbf{BD} and \textbf{DN} degradation models demonstrate the effectiveness and robustness of our RDN model. 
\vspace{-3mm}

\begin{figure}[htbp]
\centering
\includegraphics[width = 84mm]{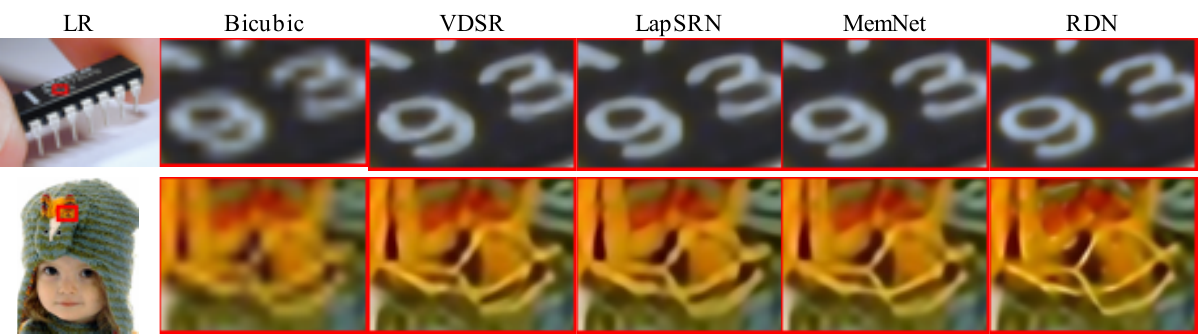}
\caption{Visual results on real-world images with scaling factor $\times4$. The two rows show SR results for images ``chip" and ``hatc'' respectively.}
\label{fig:visual_realdata} 
\vspace{-5mm} 
\end{figure}
\vspace{-1mm}
\subsection{Super-Resolving Real-World Images}
\label{subsec:realworld}
\vspace{-1mm}
We also conduct SR experiments on two representative real-world images, ``chip" (with 244$\times$200 pixels) and ``hatc" (with 133$\times$174 pixels)~\cite{zhang2017collaborative}. In this case, the original HR images are not available and the degradation model is unknown either. We compare our RND with VDSR~\cite{kim2016accurate}, LapSRN~\cite{lai2017deep}, and MemNet~\cite{tai2017memnet}. As shown in Fig.~\ref{fig:visual_realdata}, our RDN recovers sharper edges and finer details than other state-of-the-art methods. These results further indicate the benefits of learning dense features from the original input image. The hierarchical features perform robustly for different or unknown degradation models.
\vspace{-3mm}
\section{Conclusions}
\vspace{-2mm}
In this paper, we proposed a very deep residual dense network (RDN) for image SR, where residual dense block (RDB) serves as the basic build module. In each RDB, the dense connections between each layers allow full usage of local layers. The local feature fusion (LFF) not only stabilizes the training wider network, but also adaptively controls the preservation of information from current and preceding RDBs. RDB further allows direct connections between the preceding RDB and each layer of current block, leading to a contiguous memory (CM) mechanism. The local residual leaning (LRL) further improves the flow of information and gradient. Moreover, we propose global feature fusion (GFF) to extract hierarchical features in the LR space. By fully using local and global features, our RDN leads to a dense feature fusion and deep supervision. We use the same RDN structure to handle three degradation models and real-world data. Extensive benchmark evaluations well demonstrate that our RDN achieves superiority over state-of-the-art methods. 
\vspace{-2mm}
\section{Acknowledgements}
\vspace{-2mm}
This research is supported in part by the NSF IIS award 1651902, ONR Young Investigator Award N00014-14-1-0484, and U.S. Army Research Office Award W911NF-17-1-0367.

{\small
\bibliographystyle{ieee}
\bibliography{SR_conf_bib}
}

\end{document}